\documentclass[11pt]{article}

\usepackage[preprint]{acl}

\usepackage{times}
\usepackage{hyperref}
\usepackage{latexsym}
\usepackage{soul} % Include the soul package for highlighting
\definecolor{TeacherColor}{HTML}{B3CDE3}
\definecolor{StudentColor}{HTML}{CCEBC5}

\usepackage[T1]{fontenc}
\usepackage[utf8]{inputenc}

\usepackage{microtype}
\usepackage[most]{tcolorbox}
\tcbset{
  compactbox/.style={
    boxsep=1pt,
    left=2pt,
    right=2pt,
    top=1pt,
    bottom=1pt,
    toptitle=1pt,
    bottomtitle=1pt,
    before skip=6pt,
    after skip=6pt
  }
}
\usepackage{url}

\usepackage{inconsolata}
\usepackage{longtable}
\usepackage[table]{xcolor} % For \rowcolor
\usepackage{graphicx}
\usepackage{subcaption}
\usepackage{amsmath}
\usepackage{amssymb}
\usepackage{pifont}
\usepackage{booktabs} % For formal tables
\usepackage[linesnumbered,ruled,vlined]{algorithm2e}
\usepackage{multirow}
\usepackage{float}
\usepackage{xcolor} % For color definitions with HTML option
\newcommand{\myparagraph}[1]{\noindent\textbf{#1}}
\definecolor{violet}{HTML}{BEBADA}

\title{Cross-lingual Matryoshka Representation Learning across Speech and Text}

\author{
  \textbf{Yaya Sy\textsuperscript{1}},
  \textbf{Dioula Doucouré\textsuperscript{2}},
  \textbf{Christophe Cerisara\textsuperscript{1}},
  \textbf{Irina Illina\textsuperscript{1}},
\\
\\
  \textsuperscript{1}LORIA, CNRS,
  \textsuperscript{2}Soynade Research,
\\
  \small{
    \textbf{Correspondence:} \href{mailto:yaya.sy@loria@fr}yaya.sy@loria.fr
  }
}

\begin{document}
\maketitle
\begin{abstract}
Speakers of under-represented languages face both a \textbf{language barrier}, as most online knowledge is in a few dominant languages, and a \textbf{modality barrier}, since information is largely text-based while many languages are primarily oral. We address this for French-Wolof by training the first bilingual speech-text Matryoshka embedding model, enabling efficient retrieval of French text from Wolof speech queries without relying on a costly ASR-translation pipelines. We introduce large-scale data curation pipelines and new benchmarks, compare modeling strategies, and show that modality fusion within a frozen text Matryoshka model performs best. Although trained only for retrieval, the model generalizes well to other tasks, such as speech intent detection, indicating the learning of general semantic representations. Finally, we analyze cost-accuracy trade-offs across Matryoshka dimensions and ranks, showing that information is concentrated only in a few components, suggesting potential for efficiency improvements. The code, the models, and the datasets will be released: \url{https://github.com/soynade-research/oolel-embed}
\end{abstract}

\section{Introduction}
\subsection{Motivation}
Access to information is limited by two major barriers. First, a \textit{language barrier}: most content is written in a few high-resource languages. Second, a \textit{modality barrier}: information retrieval systems assume text-based queries, while many under-represented languages are primarily oral. Traditional cascaded ASR-translation pipelines are costly and suffer from error propagation. We address this by training cross-lingual speech-text Matryoshka representation models, enabling direct retrieval of text documents from speech queries with flexible accuracy-cost trade-offs.

We focus on Wolof-French as a representative and socially grounded case study. Wolof is primarily spoken in Senegal and is mainly oral. Due to colonial history, administrative, educational, and informational content relevant to Wolof speakers is accessible in French, yet many Wolof speakers have limited French literacy. This creates a critical mismatch: information for Wolof speakers exists in French text, while their natural query modality is Wolof speech, making cross-lingual speech-based retrieval a practical solution.

\subsection{Related Works}
\myparagraph{Speech Language Models} extend LLMs with speech understanding by integrating a speech encoder \citep{llamaspeech, pareras2025revisitingdirectspeechtotexttranslation, slm_methods}. We similarly add speech capabilities to embedding LLMs, which remains underexplored compared to generative LLMs.
\\
\\
\myparagraph{Matryoshka Representation Learning (MRL)} \citep{MRL} reduces deployment costs by learning representations at multiple dimensions simultaneously, enabling flexible dimension choice at inference. This has been generalized to vision-language and audio-visual LLMs \citep{hu2024Matryoshkaquerytransformerlarge, cai2024Matryoshkamultimodalmodels, cappellazzo2025adaptiveaudiovisualspeechrecognition} for sequence compression at different granularities. The main novelty of our work is we extend this to cross-lingual speech-text retrieval. We analyze performance-cost trade-offs of matryoshkas and their ranks, providing key findings on how the information is represented by the different dimensions.
\\
\\
\myparagraph{Multilingual Representation Models} \citep{xlmr, mbert, mbart} demonstrate strong cross-lingual transfer for low-resource languages but are generally not designed for retrieval. Recent work \citep{nllbllm2vec} integrates a massively multilingual machine translation encoder such as NLLB into an LLM to produce cross-lingual representations for over 200 languages. We show that a model 10x smaller, augmented with speech capability and trained mostly on supervised synthetic bilingual data, not only supports retrieval but also learns robust representations that transfer to non-retrieval tasks.
\\
\\
\myparagraph{CLIP-style Architectures} \citep{clip} use modality-specific encoders with contrastive objectives. CLAP \citep{clap} extends this to audio-text. We show these Dual-encoder models work for simple tasks like transcription retrieval but struggle with semantically demanding cross-lingual speech-to-document retrieval.
\section{Datasets}
\label{sec:dataset}
To address the scarcity of Wolof speech-French text retrieval data, we primarily rely on synthetic documents. We first describe the training data collection pipelines for both text-only and speech-text data, then introduce the evaluation datasets. In this work, \textit{query} is any speech or text used as a query for retrieval, whether it is a question or not.

\begin{figure}
    \includegraphics[width=\linewidth]{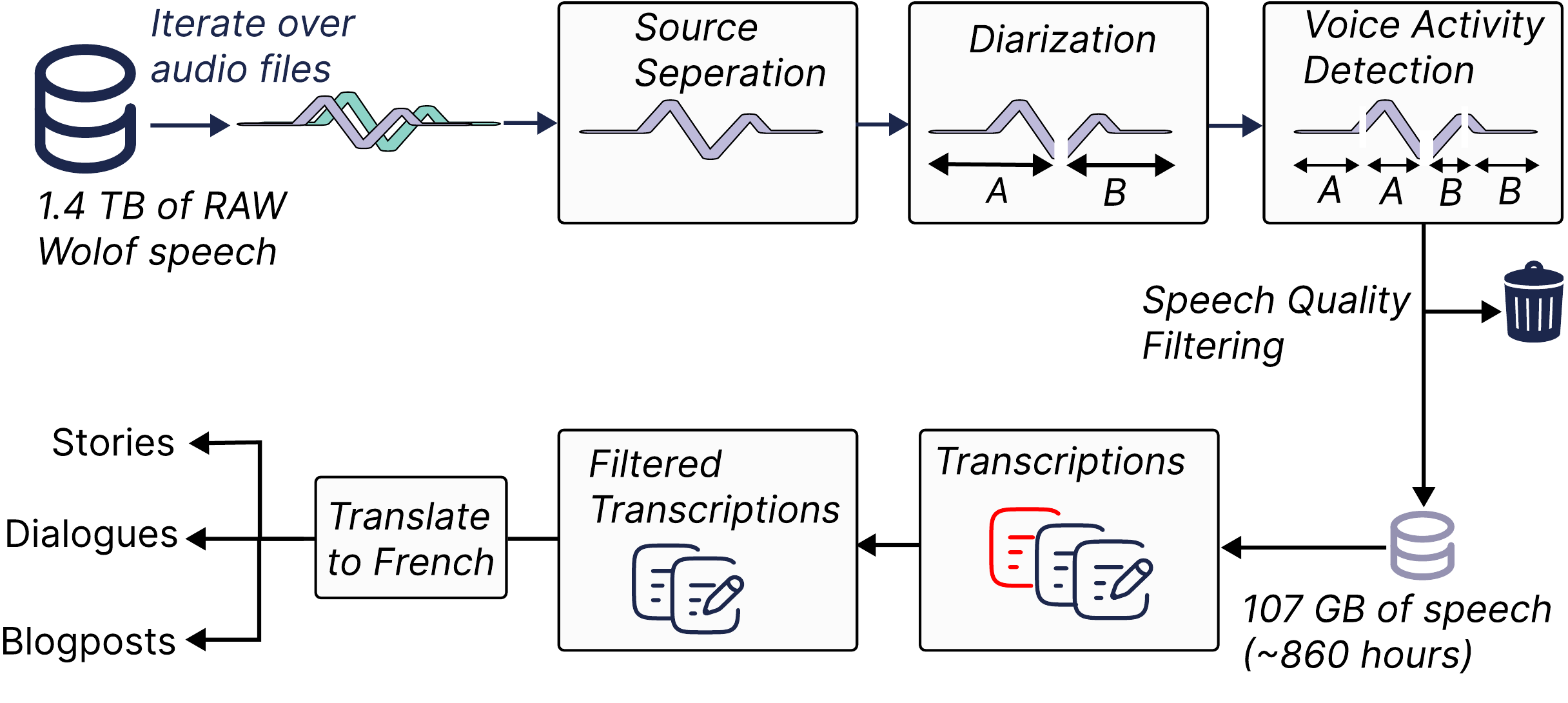}
    \caption{\textbf{Speech data pipeline}. Raw data is filtered via \textit{Source Separation}, \textit{Diarization}, \textit{VAD}, and \textit{Quality Filtering}. The resulting speech is transcribed, bad transcriptions (in {\color{red} red}) are filtered, then used to generate French \textit{story, dialog} and \textit{blogpost} documents.}
    \label{fig:pipeline}
\end{figure}
\subsection{Training Datasets}
\myparagraph{Text-only Training Dataset.} We curate data from three sources: (1) French mMARCO \citep{mmarco} with queries translated to Wolof, (2) Senegalese French webpages, where we synthetically generate French queries then translate them to Wolof. We augment the training dataset with (3) French Question-Answering datasets, where the questions are translated to Wolof. We add Wolof-French translation pairs for cross-lingual transfer. The total text-only training dataset yields 1,176,908 query (Wolof) and document (French) pairs, corresponding to 593,284,495 of document tokens.
\\
\\
\myparagraph{Collecting Wolof Speech Queries} We collect speech data by manually sourcing natural and spontaneous speech (podcasts, radio programs, etc.) by web browsing and listening to audio. We exclude read content like audiobooks. The scraped data over these sources yields 1.4TB of speech, but requires pre-processing for training. We evaluate multiple filtering pipelines and select one that effectively removes long silences, excessive backchannels, and unintelligible overlap. The filtering pipeline is presented in Figure \ref{fig:pipeline}. We iterate through the raw 1.4TB of audio and apply, in order: \textit{Source Separation}\footnote{we used \url{https://huggingface.co/seanghay/uvr_models/blob/main/UVR-MDX-NET-Inst_HQ_3.onnx}} to isolate speech from other sounds, \textit{Diarization} from \texttt{pyannote} \citep{Plaquet23, Bredin23} to split audio by speaker, and \textit{Voice Activity Detection} from \texttt{Silero-VAD} \citep{Silero-VAD} to retain segments with speech. Finally, we keep only utterances between 3-30 seconds with a DNSMOS \cite{dnsmos} quality score above 3.2, resulting in 860 hours of high-quality Wolof speech queries.
\\
\\
\myparagraph{Synthetic French Documents.} Starting from Wolof speech queries, we synthesize French documents. We first transcribe the Wolof speech using a Wolof Speech Language Model \citep{sy2025speechlanguagemodelsunderrepresented}, then filter transcriptions based on perplexity and lexical diversity (unique-to-total word ratio), retaining approximately one quarter of the original 860 hours of Wolof speech-text pairs. The filtered Wolof transcriptions are translated into French, and these translations are used to generate three types of synthetic documents with Gemini-2.5-Flash: stories, dialogues, and web blogposts.
\\
\\
\myparagraph{Instruction-Following Dataset.} In real-world scenarios, databases are heterogeneous: the same input may correspond to multiple target documents. For example, for the same Wolof speech query, it is possible to retrieve in the database either the corresponding Wolof transcription or the French document. Instruction-following embedding models \cite{peng-etal-2024-answer} address this by appending a task description to the query. Accordingly, we train a multitask embedding model that can be prompted at inference time, rather than optimizing solely for French document retrieval. In addition to document retrieval, we include \textit{speech translation retrieval} (retrieving French translations from Wolof speech) and \textit{transcription retrieval} (retrieving Wolof transcriptions from speech), which is useful for applications such as keyword spotting. Appendix \ref{sec:asr} details the dataset for these additional tasks.

\subsection{Test Datasets}
To evaluate the models, we introduce two benchmarks for Wolof to French document retrieval. The first is derived from SIB-Fleurs, a benchmark based on Fleurs test-split. The second benchmark is based on the Kallaama test-split~\cite{kallama}. We detail next both benchmarks.
\begin{tcolorbox}[compactbox,colback=violet!15!white,colframe=violet!100!white,title={\color{black} \textbf{Artefact 1}: Evaluation Datasets for French document retrieval from Wolof queries.}]
We introduce \textit{Kallaama-Retrieval-Eval} and \textit{Fleurs-Retrieval-Eval}, two datasets for evaluating French document retrieval from Wolof text/speech queries.
\end{tcolorbox}
\myparagraph{Fleurs-Retrieval-Eval} Because most of our training data is synthetic, we evaluate on a fully natural benchmark. We start from SIB-Fleurs \citep{schmidt2025fleursslumassivelymultilingualbenchmark}, a multilingual speech topic classification dataset based on the Fleurs test split \citep{fleurs}. Using the French translations of the Wolof speech, we retrieve the most relevant French documents from the web, and manually filter out cases where the scraper fails or returns irrelevant results. The retrieved documents are used as-is, without cleaning or shortening, to reflect real-world usage. The resulting \textit{Fleurs-Retrieval-Eval} dataset contains 166 pairs of Wolof speech and corresponding French text documents.
\\
\\
\myparagraph{Kallaama-Retrieval-Eval} After manual verification and listening to the Fleurs audio, we found that the Wolof speech is unnatural, as speakers are hesitant and speak quietly. We will empirically support and explain this later in the analysis Section \ref{sec:analysis}. We propose another evaluation dataset that relies on the test split of Kallaama~\cite{kallama}, a dataset where native Wolof speakers converse naturally, and the speech is transcribed by professionals. From the test split, we select the 150 longest speech queries that do not exceed 30 seconds, ensuring information-rich queries while respecting the input length constraints of modern speech models. We translate the transcriptions into French and use Gemini-2.5-Pro to synthetically generate three document types: \textit{dialogues}, \textit{blogposts}, and \textit{stories}. The resulting evaluation set contains 150 Wolof speech and text queries, each paired with three corresponding French documents.

\section{Modeling}
\label{sec:model}
\begin{figure}[ht]
  \includegraphics[width=\linewidth]{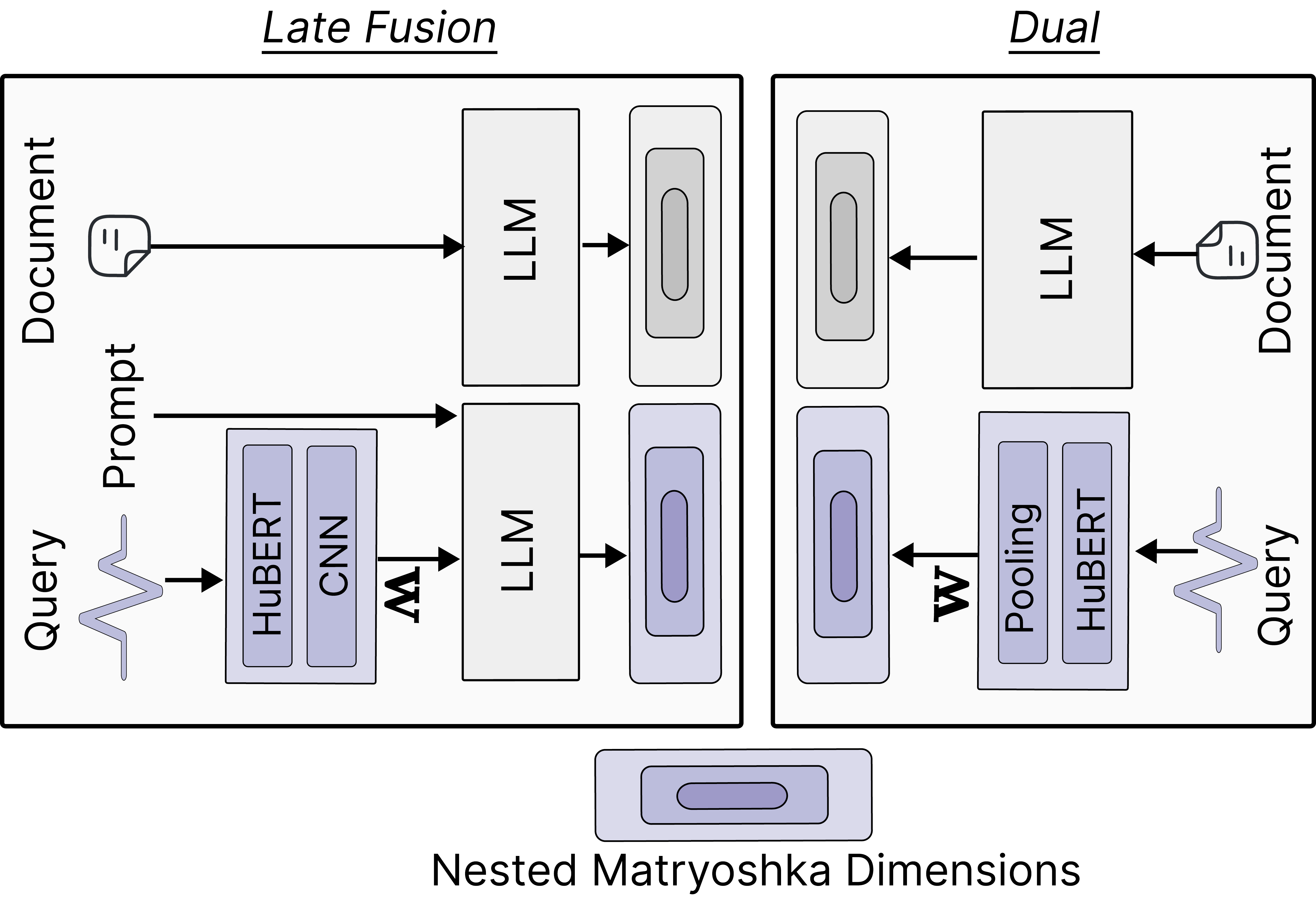}
  \caption{\textbf{Late-Fusion vs. Dual architectures}. In the \textit{Late-Fusion} approach, speech is encoded with HuBERT, then the sequence is downsampled by a CNN (x2), projected to the LLM embedding with a matrix $\textbf{W}$, concatenated with the prompt token embeddings, and the whole is forwarded to the Matryoshka embedding LLM. In the \textit{Dual} architecture, HuBERT features are pooled at the sequence level using an attention-based pooler, then projected with dimension-specific $\textbf{W}$ matrices to obtain Matryoshka embeddings. In both architectures, text documents are embedded using the text-only Matryoshka embedding LLM.}
  \label{fig:models}
\end{figure}
\noindent To enable flexible choice of the dimensions at inference-time, Matryoshka Representation Learning~\cite{MRL} optimizes a joint retrieval loss across different dimensions. Given the training loss $\mathcal{L}$ and the Matryoshka dimensions $\mathcal{M} = \{d_{1}, d_{2}, \cdots, d_{m}\}$ where $d_{1} < d_{2} < \cdots < d_{m}$, the model optimizes this joint loss: 
$$
\mathcal{L}_{MRL} = \sum\limits_{m \in \mathcal{M}} \mathcal{L}(\textbf{Q}_{:m}, \textbf{D}_{:m})
$$
where $\textbf{Q}$ is the query embedding matrix, $\textbf{D}$ is the document embedding matrix, and $:m$ refers to the PyTorch slicing operator. We use InfoNCE \citep{infonce} as retrieval loss, with in-batch negative, meaning for each query-document pair, the negatives are all other documents in the batch. Next, we detail the text-only and multimodal models: \textbf{Late-Fusion} and \textbf{Dual} encoders (see Figure \ref{fig:models}).

\subsection{Text-only Embedding}
We use Qwen3-0.6-Embedding, an MRL Embedding LLM that can represent text in dimensions 32, 64, 128, 256, 512, and 1024. Since the training of Matryoshka is costly when there are too many dimensions, due to the loss summation, we only use dimensions 128, 256, 512, and 1024. We first fine-tune Qwen3-0.6-Embedding on the text-only data presented previously, so the model learns cross-lingual representations between Wolof and French. This model is trained with in-batch InfoNCE loss. We next introduce and study two approaches for integrating the speech modality within this text-only model.
\begin{tcolorbox}[compactbox,colback=violet!15!white,colframe=violet!100!white,title={\color{black} \textbf{Artefact 2}: Cross-lingual speech and text representation models for Wolof and French}]
We introduce a series of cross-lingual Matryoshka retrieval models for Wolof and French. This includes both text-only and speech-text representation models.
\end{tcolorbox}
\subsection{Late Modality Fusion}
In \textit{Late-Fusion} \cite{llava, llamaspeech} is a simple and efficient approach for integrating vision or speech capabilities into pre-trained LLMs. In this approach, the speech features from the speech encoder and text token embeddings from the embedding layer are produced separately and then concatenated before being fed into the pretrained language model. The \textit{fusion} is late because the two modalities only interact within the language model, as opposed to \textit{early-fusion}, where the modalities are combined before. For the speech encoder, we use a continued-pretrained Wolof HuBERT model \citep{sy2025speechlanguagemodelsunderrepresented}. Figure \ref{fig:models} illustrates this approach. Wolof query speech is forwarded through HuBERT; instead of using only the final layer, features from all 12 HuBERT layers are concatenated, passed through a CNN to reduce the sequence length, and projected to the LLM embedding space using $\mathbf{W}$. The resulting speech embeddings are concatenated at the sequence level with the prompt token embedding and forwarded to the LLM. The document is embedded separately by the LLM. The model is trained end-to-end on the Instruction-Following Speech-Text dataset using a joint InfoNCE loss over the different Matryoshka dimensions. Only the CNN and $\mathbf{W}$ are trained, while HuBERT and the LLM remain frozen.

\subsection{Dual Modality}
An alternative is a Dual-encoder architecture, similar to CLIP \citep{clip} or CLAP \citep{clap}, illustrated on the right of Figure \ref{fig:models}. The pipeline is similar to \textit{Late-Fusion}, except that speech features are not forwarded to the LLM. In this approach, a pooling function is required to map the speech sequence vectors to a single vector. This is unnecessary in Late-Fusion, which naturally uses the pooling mechanism of Qwen3-0.6B-Embedding, namely the final token representation. For \textit{Dual} architectures, we use \textit{attention-based} pooling with a learnable query. We define a pooling query parameter $\textbf{q} \in \mathbb{R}^{1\times d}$, and given the HuBERT features $\textbf{X} \in \mathbb{R}^{s\times d}$, the pooling is defined as $\text {softmax}(\frac{\textbf{q} \textbf{X}^{T}}{\sqrt{d}}) \textbf{X}$

Compared to the Late-Fusion approach, this Dual-encoder architecture has notable limitations. First, it is inherently less expressive, as speech queries are represented without the depth of contextualization provided by the pre-trained language model's transformer layers. Second, the Dual approach is incompatible with \textit{task prompting}, since the speech query is encoded solely by the speech modules, which cannot process text prompts. So, to provide more expressivity during training, we unfreeze the HuBERT, and in place of the simple slicing, we introduce different trainable linear projections for each Matryoshka dimension. We study two different training objectives for the Dual architectures.
\\
\\
\myparagraph{Dual - \textit{Retrieval}.} This first approach trains the Dual architecture with in-batch InfoNCE retrieval loss. During training, the text LLM is frozen, contrary to the speech modules which are unfrozen.
\\
\\
\myparagraph{Dual - \textit{Query Alignment}.} We found that Dual approaches trained on document retrieval fail to converge well, likely because the 90M-parameter HuBERT model is too small. We study a simpler alternative: aligning Wolof speech query representations with their transcriptions via distillation. Since French document retrieval from Wolof text queries works well, aligning the representation speech queries with their transcriptions should enable Wolof speech to French document retrieval without direct training of speech to document retrieval. We use a distillation loss consisting of a joint loss of cosine similarity loss (\texttt{CosineSimilarity} in PyTorch) and $l_{1}$ loss (\texttt{L1Loss} in PyTorch).

\section{Experiments}
\myparagraph{Setup.}
Using the Sentence-Transformers library, we implement the four models: \textit{text-only}, \textit{Late-Fusion}, \textit{Dual-retrieval}, and \textit{Dual-query alignment}. All models are trained for 1 epoch with batch size 16, maximum length 2048, and learning rate $3\cdot 10^{-4}$. The \textit{text-only} and \textit{Late-Fusion} models are trained on their respective instruction datasets. The Dual-encoder models use the same data as \textit{Late-Fusion} but without prompts, since the query is only processed by the speech encoder that cannot process texts.
\\

\myparagraph{Compute.}
We used DeepSpeed for distributed training on 8xA100 GPUs to fine-tune the Qwen3-0.6B text embedding model, with a training time of ~2 hours. Assuming a rental cost of \$1.40 per hour per A100, this corresponds to an estimated cost of \$22.4 for this stage. The late-fusion and dual models are cheaper to train, as the text embedding model, which is the biggest part, is frozen and only the small speech aligner (in the late-fusion approach) or the speech encoder (in the dual approach) is updated. These models were trained using a single H100 GPU for approximately 8 hours.

\begin{table*}[t]
    \resizebox{\textwidth}{!}{%
    \begin{tabular}{ccccccccccc}
        \toprule
        \multirow{2}*{\textbf{Approach}} & \multicolumn{2}{c}{\textbf{dim=4096}} & \multicolumn{2}{c}{\textbf{dim=1024}} & \multicolumn{2}{c}{\textbf{dim=512}} & \multicolumn{2}{c}{\textbf{dim=256}} & \multicolumn{2}{c}{\textbf{dim=128}} \\
        \cmidrule(lr){2-3} \cmidrule(lr){4-5} \cmidrule(lr){6-7} \cmidrule(lr){8-9} \cmidrule(lr){10-11}
        & \textbf{nDCG@5} & \textbf{nDCG@10} & \textbf{nDCG@5} & \textbf{nDCG@10} & \textbf{nDCG@5} & \textbf{nDCG@10} & \textbf{nDCG@5} & \textbf{nDCG@10} & \textbf{nDCG@5} & \textbf{nDCG@10} \\
        \midrule
        NLLB-LLM2Vec & 57.98 & 61.53 & -- & -- & -- & -- & -- & -- & -- & -- \\
        \midrule
        \rowcolor{violet!45} Late-Fusion & -- & -- & \textbf{69.85} & \textbf{74.49} & 66.86 & 71.04 & 61.58 & 67.05 & 56.13 & 62.30 \\
        Pipelined & -- & -- & 57.09 & 62.82 & 54.07 & 59.14 & 45.60 & 51.56 & 41.21 & 47.34 \\
        
        Dual -- \textit{Retrieval} & -- & -- & 46.96 & 53.70 & 45.27 & 52.48 & 41.97 & 49.78 & 38.47 & 44.40 \\
        
        Dual -- \textit{Query Alignment} & -- & -- & 41.56 & 47.42 & 41.02 & 46.76 & 38.24 & 44.18 & 35.47 & 40.47 \\
        \bottomrule
    \end{tabular}}
    \caption{nDCG@5 and nDCG@10 document retrieval results on \textbf{Kallaama-Retrieval-Eval} for the different approaches across dimensions. NLLB-LLM2Vec is not Matryoshka and uses the full model dimension (4096), while other approaches use Matryoshka dimensions. The most performant speech-based approach is colorized and the best scores are in bold.}
    \label{tab:embedding_results_kallama}
\end{table*}
\begin{table*}[t]
    \resizebox{\textwidth}{!}{%
    \begin{tabular}{ccccccccccc}
        \toprule
        \multirow{2}*{\textbf{Approach}} & \multicolumn{2}{c}{\textbf{dim=4096}} & \multicolumn{2}{c}{\textbf{dim=1024}} & \multicolumn{2}{c}{\textbf{dim=512}} & \multicolumn{2}{c}{\textbf{dim=256}} & \multicolumn{2}{c}{\textbf{dim=128}} \\
        \cmidrule(lr){2-3} \cmidrule(lr){4-5} \cmidrule(lr){6-7} \cmidrule(lr){8-9} \cmidrule(lr){10-11}
        & \textbf{nDCG@5} & \textbf{nDCG@10} & \textbf{nDCG@5} & \textbf{nDCG@10} & \textbf{nDCG@5} & \textbf{nDCG@10} & \textbf{nDCG@5} & \textbf{nDCG@10} & \textbf{nDCG@5} & \textbf{nDCG@10} \\
        \midrule
        NLLB-LLM2Vec & 55.98 & 59.43 & -- & -- & -- & -- & -- & -- & -- & -- \\
        \midrule
        \rowcolor{violet!45} Late-Fusion & -- & -- & \textbf{57.89} & \textbf{61.19} & 55.03 & 58.90 & 50.87 & 54.43 & 41.56 & 46.60 \\
        
        Dual -- \textit{Retrieval} & -- & -- & 41.28 & 45.54 & 40.18 & 45.06 & 39.84 & 45.17 & 39.24 & 44.34 \\
        
        Dual -- \textit{Query Alignment} & -- & -- & 38.07 & 41.82 & 37.33 & 41.69 & 38.04 & 41.29 & 34.46 & 38.53 \\
        \bottomrule
    \end{tabular}}
    \caption{nDCG@5 and nDCG@10 document retrieval results on \textbf{Fleurs-Retrieval-Eval} for the different approaches across dimensions. NLLB-LLM2Vec uses the full embedding dimension (4096), while other approaches use Matryoshka dimensions. The most performant speech-based approach is colorized and the best scores are in bold.}
    \label{tab:embedding_results_fleurs}
\end{table*}
\subsection{Experiment 1: Retrieval Tasks}
\myparagraph{Evaluation.} We evaluate the trained models on Kallaama-Retrieval-Eval and Fleurs-Retrieval-Eval using nDCG, a standard metric to evaluate the ranking of recommender systems. 
nDCG@$k$ measures the ranking quality of the top $k$ retrieved documents by comparing their graded relevance to an ideal ranking, with higher scores indicating better alignment with the ground truth.
We implement the evaluation script using \texttt{InformationRetrievalEvaluator} class from Sentence-Transformers.
\\
\\
\myparagraph{Baseline.} We compare our speech–text retrieval models with NLLB-LLM2Vec \citep{nllbllm2vec}\footnote{{\tiny \url{https://huggingface.co/fdschmidt93/NLLB-LLM2Vec-Meta-Llama-31-8B-Instruct-mntp-unsup-simcse}}}
, a massively multilingual text-only encoder built on the NLLB machine translation model and LLaMA3-8B, trained via self-distillation and supporting over 200 languages, including Wolof. NLLB-LLM2Vec has over 8B parameters—about 10x more than our models, and produces fixed 4096-dimensional embeddings, whereas our approach outputs Matryoshka embeddings ranging from 128 to 1024 dimensions. We also include a \textit{pipelined} baseline, where speech is first transcribed and the transcription is then used as a text query with our text-only model.
\\
\\
\myparagraph{Results.} Table \ref{tab:embedding_results_kallama} presents retrieval results on the \textit{Kallaama-Retrieval-Eval} dataset. The \textit{Late-Fusion} model outperforms the text-only NLLB-LLM2Vec baseline despite being 10× smaller and using 4x fewer embedding dimensions. It also surpasses the pipelined approach, which is affected by transcription error propagation. These results indicate that directly leveraging speech features through Late-Fusion reduces error accumulation and yields more effective retrieval. The Dual architecture performs poorly overall, although training with a \textit{retrieval} objective yields better results than \textit{Query Alignment}. Table \ref{tab:embedding_results_fleurs} also reports results on the \textit{Fleurs-Retrieval-Eval} dataset. While Late-Fusion still performs best, the performance gap across Matryoshka dimensions is larger. As we show later, this is due to the lower speech quality in Fleurs, indicating that higher Matryoshka dimensions are preferable in low-quality speech settings.
\begin{tcolorbox}[compactbox,colback=violet!15!white,colframe=violet!100!white,title={\color{black} \textbf{Finding 1}: \textit{Late-Fusion} overcomes \textit{Dual} architectures for fast cross-modal adaptation}]
Compared to \textit{Dual} architectures, \textit{Late-Fusion} shows better cross-modal generalization while having fewer trainable parameters. We conclude that \textit{Late-Fusion} shows better cross-modal generalization.
\end{tcolorbox}
\subsection{Experiment 2: Speech Keyword Spotting}
\myparagraph{Evaluation.} We evaluate the models on speech keyword spotting using the test split of Urban Bus \citep{urbanbus}, a dataset of common places in Dakar city. The task is to detect the station or common place pronounced by the user. This is exactly transcription retrieval, and our models have learned such tasks during training.

\myparagraph{Results.} Table \ref{tab:kws} reports Keyword Spotting Recall and F1-Scores across Matryoshka dimensions. As for retrieval tasks, the \textit{Late-Fusion} approach outperforms the \textit{Dual} architecture at all dimensions. Compared to the retrieval results in Tables \ref{tab:embedding_results_kallama} and \ref{tab:embedding_results_fleurs}, the performance of Dual architectures is higher overall, suggesting that for less semantically demanding tasks such as transcription retrieval, the Dual architecture can be a viable solution. Training the Dual architecture with the \textit{Query Alignment} objective yields better performance than the \textit{retrieval} objective.
\begin{tcolorbox}[compactbox,colback=violet!15!white,colframe=violet!100!white,title={\color{black} \textbf{Finding 2}: \textit{Dual} models perform well on tasks that are not semantically demanding.}]
\textit{Dual} architectures perform poorly on speech-to-document retrieval, but show good results on transcription retrieval tasks, which don't require deep semantic understanding.
\end{tcolorbox}
\subsection{Experiment 3: Unseen Task}
\label{sec:unseen_task}

\begin{table*}[ht]
    \begin{minipage}[c]{.62\textwidth}
        \fontsize{9pt}{9pt}\selectfont
        %\vspace{-20.5em}
        \centering
        \addtolength{\tabcolsep}{-0.3em}
        \begin{tabular}{@{}p{2.2cm}lcccc@{}}
            \toprule
            \multicolumn{2}{c}{\textbf{Approach}} & \textbf{dim=1024} & \textbf{dim=512} & \textbf{dim=256} & \textbf{dim=128} \\
            \midrule
            \multirow{2}{2.2cm}{Late-Fusion} & \cellcolor{violet!45}\textbf{F1-Score} & \cellcolor{violet!45}88.79 & \cellcolor{violet!45}84.85 & \cellcolor{violet!45}79.80 & \cellcolor{violet!45}76.86 \\
            & \cellcolor{violet!45}\textbf{Recall} & \cellcolor{violet!45}89.79 & \cellcolor{violet!45}86.49 & \cellcolor{violet!45}81.98 & \cellcolor{violet!45}79.88 \\
            \addlinespace
            \multirow{2}{2.2cm}{Dual - \textit{Retrieval}} & \textbf{F1-Score} & 68.64 & 58.33 & 50.86 & 44.59 \\
            & \textbf{Recall} & 71.17 & 62.76 & 56.76 & 52.25 \\
            \addlinespace
            \multirow{2}{2.2cm}{Dual - \textit{Query Alignment}} & \textbf{F1-Score} & 76.07 & 67.67 & 67.67 & 65.22 \\
            & \textbf{Recall} & 77.78 & 70.87 & 70.87 & 68.47 \\
            \bottomrule
        \end{tabular}
        \caption{\textbf{Keyword Spotting} (Urban Bus) Performance Comparison across different embedding dimensions.}
        \label{tab:kws}
    \end{minipage}
    \hfill
    \begin{minipage}[c]{.35\textwidth}
        \fontsize{9pt}{9pt}\selectfont
        \centering
        \addtolength{\tabcolsep}{-0.1em}
        \begin{tabular}{ccc}
            \toprule
            \textbf{\textit{n}-shot} & \textbf{F1-Score} & \textbf{Recall} \\
            \midrule
             0 & 44.79 & 50.64 \\
              1 & 56.29 & 58.50 \\
              2 & 60.40 & 61.93 \\
              4 & 87.18 & 88.46 \\
              8 & 92.54 & 92.51 \\
              16 & 96.11 & 96.10 \\
            \bottomrule
        \end{tabular}
        \captionof{table}{\textit{n}-shot Performance for \textbf{Speech Intent Detection} at dimension 1024.}
        \label{tab:fewshot1024}
    \end{minipage}
\end{table*}

So far, we have evaluated performance on tasks seen during training. We now assess the models’ generalization to an unseen task using a speech intent detection task in a few-shot setting.

\subsubsection{Data}
We use WolBanking77 \citep{wolbanking}, an intent detection dataset with both speech and text queries, where text queries are transcriptions of the speech. The task is to classify each speech request into one of 10 intents.

\subsubsection{Method}
We evaluate the generalization ability of the \textit{Late-Fusion} model in zero-shot and few-shot settings. In \textbf{the zero-shot} setting, the model encodes the intent labels and the user's speech request separately, and classifies the intent of the speech as the label with the highest similarity score. In the \textbf{n-shot} setting, the model is fine-tuned on \textit{n} examples per class in the training set. Following SetFit \citep{setfit}, the few-shot learning is performed in two stages.
\\
\\
\myparagraph{Stage 1.} In this stage, the model learns better representations of the instances of each label. The model is fine-tuned on pairs of positive and negative examples drawn from the few-shot training dataset. This is achieved by pairing each example of a class (positive) with all other examples of different classes in the dataset (negatives). We then perform contrastive training of the \textit{Late-Fusion} model on the positive-negative pairs, varying the \textit{n}-shot setting in $\{1, 2, 4, 8, 16\}$.
\\
\\
\myparagraph{Stage 2.} This second stage fine-tunes a classifier head on top of the frozen model from Stage 2, using standard multiclass logistic regression.
\subsection{Results}
\begin{figure*}[ht]
    \begin{minipage}[t]{0.32\linewidth}
        \begin{center}
            \includegraphics[width=\textwidth]{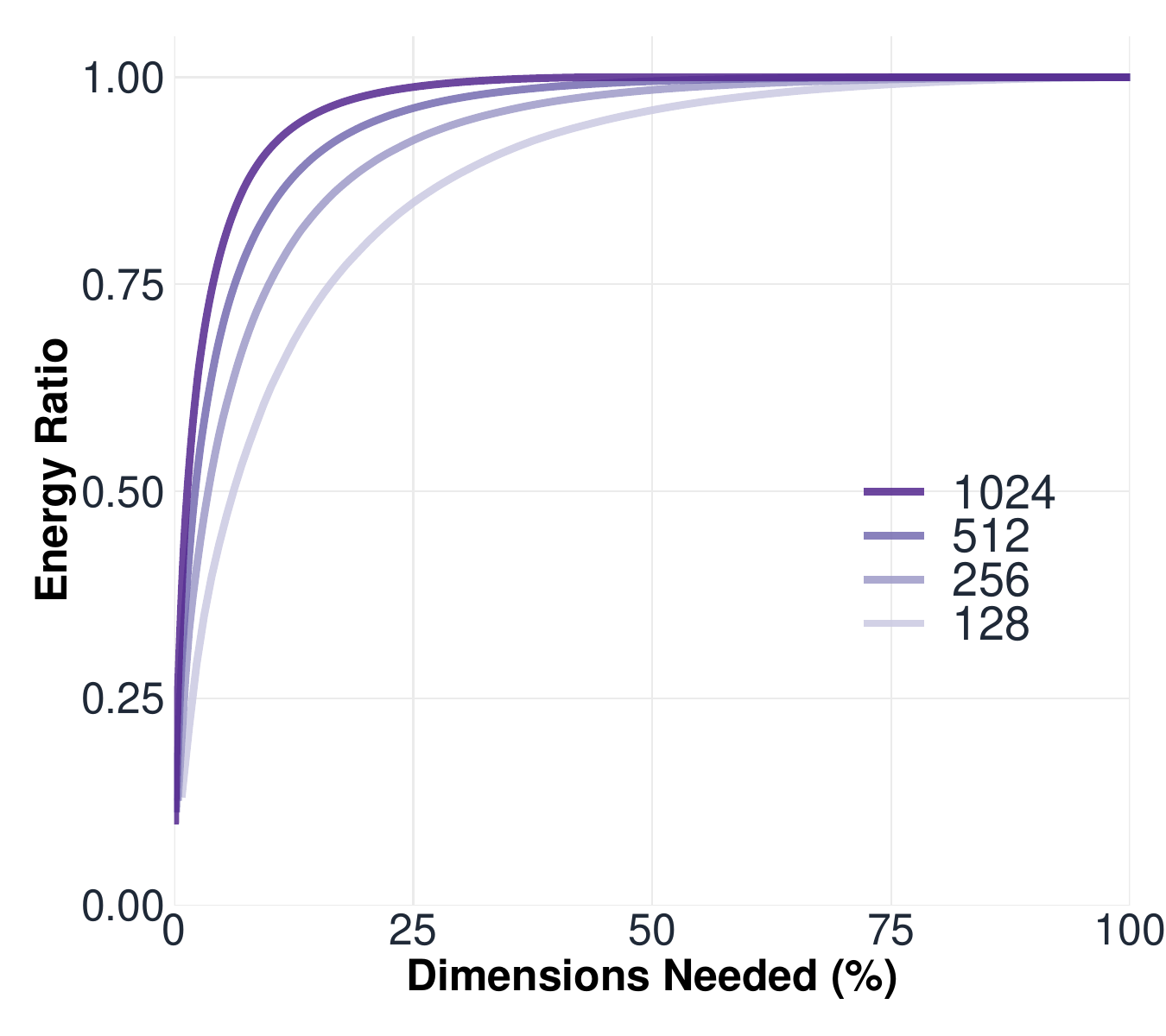}
            \caption{Percentage of dimensions needed to represent a given energy ratio.}
            \label{fig:energy_evolution}
        \end{center}
    \end{minipage}
    \hfill
    \begin{minipage}[t]{0.32\linewidth}
        \centering
        \includegraphics[width=\textwidth]{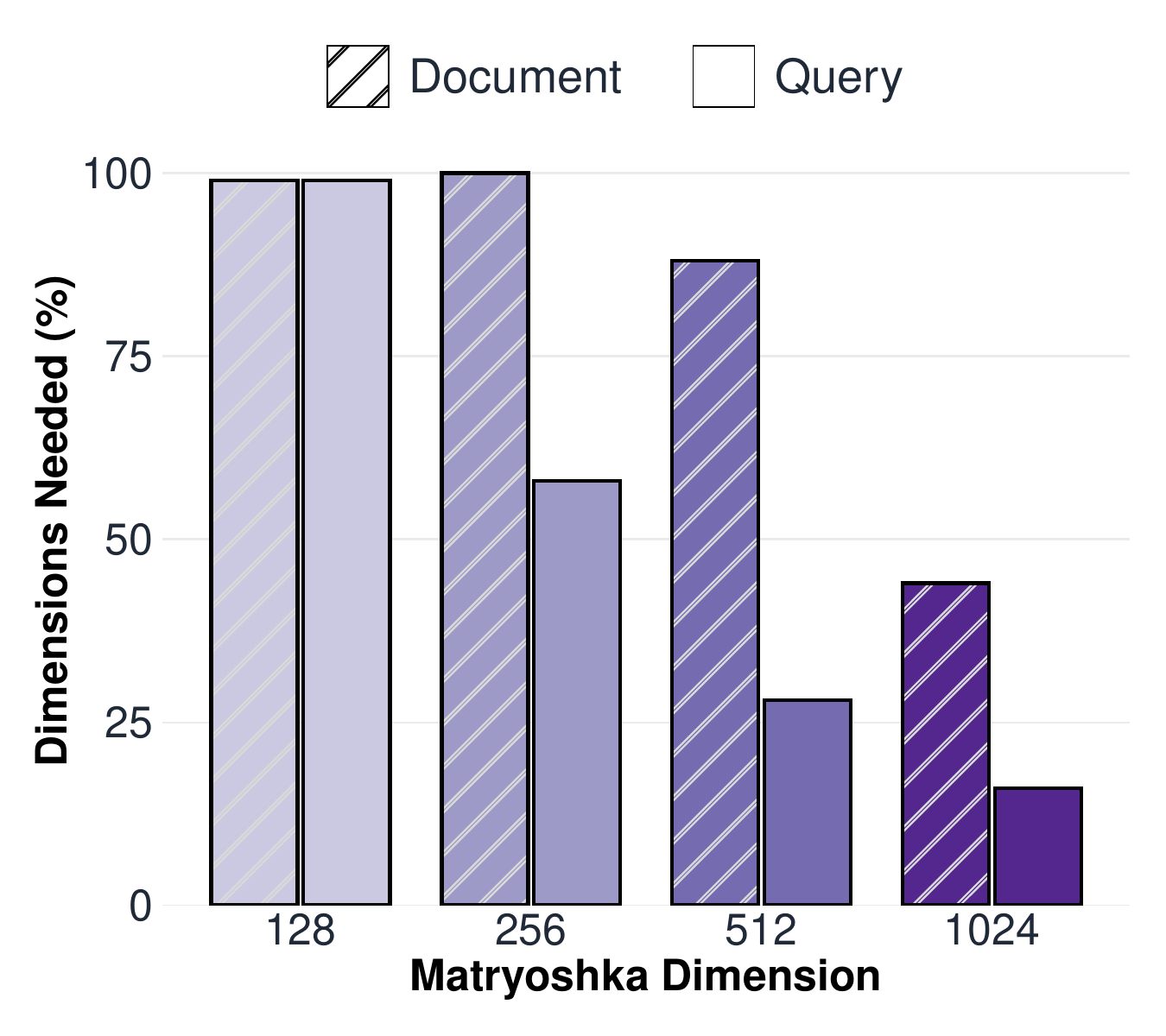}
        \caption{Percentage of dimensions needed to represent the full energy}
        \label{fig:energy_100p}
    \end{minipage}
    \hfill
    \begin{minipage}[t]{0.32\linewidth}
        \centering
        \includegraphics[width=\textwidth]{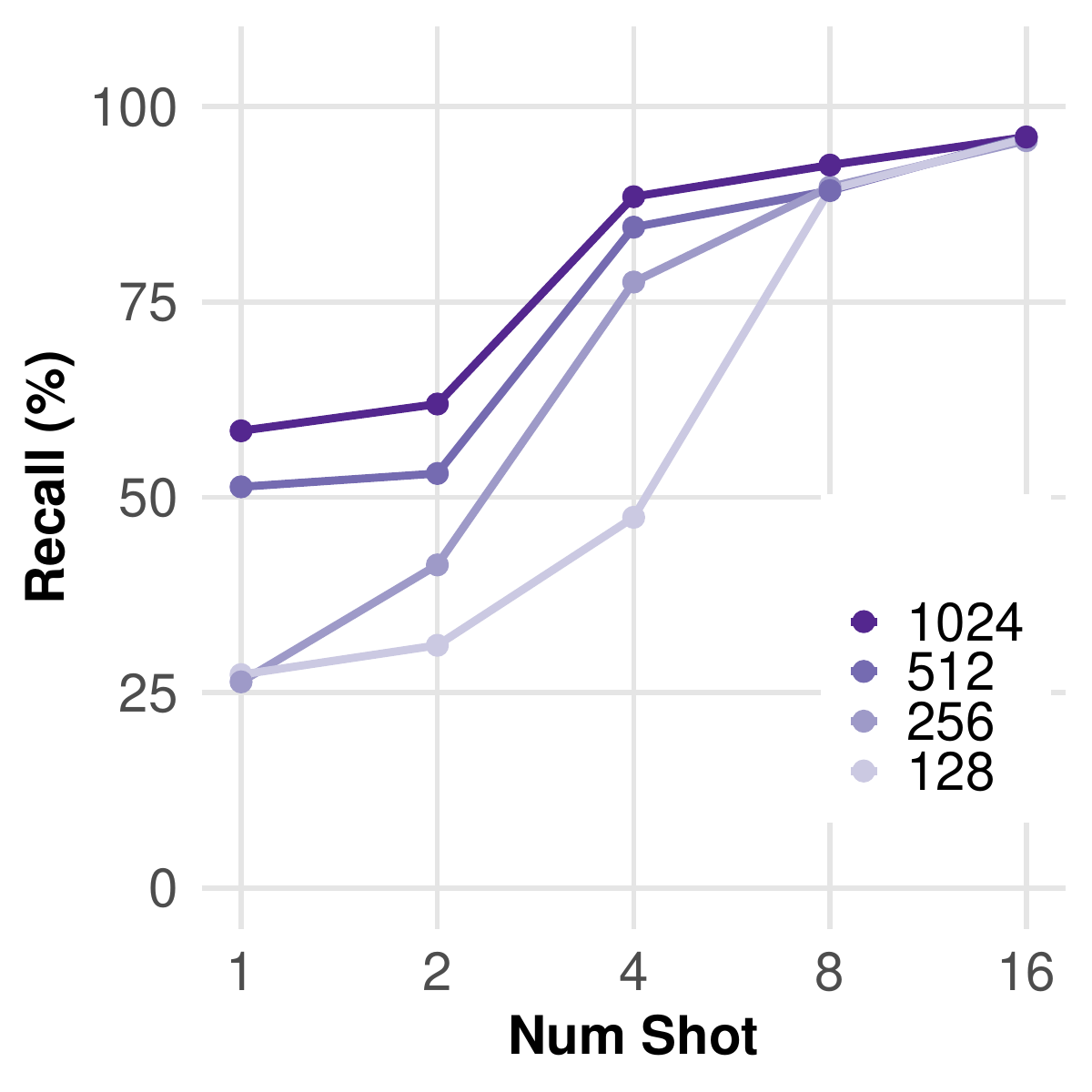}
        \caption{How the recall improves across Matryoshka dimension as a function of \textbf{few-shot examples}.}
        \label{fig:sid_dimensions}
    \end{minipage}
\end{figure*}

\myparagraph{Performance of Speech Intent Detection at 1024 dimension.} Table \ref{tab:fewshot1024} shows the results for a Matryoshka Dimension of 1024. The 0-shot F1-Score is already better than random, which is approximatively 10\%. Adding a few more examples greatly improves overall performance, achieving an F1-Score of 96.11 with 16-shot training examples.
\\
\\
\myparagraph{Performance of Speech Intent Detection across dimensions.} Figure \ref{fig:sid_dimensions} shows the evolution of Speech Intent Detection performance for different Matryoshka dimensions as the number of shots increases. While higher dimensions perform better with fewer examples, lower dimensions catch up when increasing the number of examples. Since, as we will show it, lower dimensions are cheaper at inference, these results suggest it is worthwhile to pay higher training costs in exchange for reduced inference costs.
\begin{tcolorbox}[compactbox,colback=violet!15!white,colframe=violet!100!white,title={\color{black} \textbf{Finding 3}: Big dimensions adapt quickly to new tasks, smaller ones need more data.}]
When only a few examples are available, big dimensions perform better. However, with more training examples, small dimensions catch up.
\end{tcolorbox}

\section{Analysis}
\label{sec:analysis}
\noindent We provide analysis, interpretations, and justifications of the results. First, we analyze the speech quality of Fleurs and Kallaama, showing that Fleurs' lower quality explains the weaker performance on the Fleurs-Retrieval-Eval. Second, we analyze how Matryoshka dimensions represent information by examining their rank. Third, we compare the deployment costs across different Matryoshka dimensions. And finally, we analyze the instruction-following capabilities of the model. All the analyses are done on the Late-Fusion model.

\subsection{On the speech quality of Fleurs}
\begin{tcolorbox}[compactbox,colback=violet!15!white,colframe=violet!100!white,title={\color{black} \textbf{Finding 4}: Wolof speech in Fleurs is unnatural.}]
Wolof is mainly an oral language. The speakers are not used to reading it, leading to more \textit{hesitations} and \textit{lower volume} in Fleurs read speech.
\end{tcolorbox}
We analyze the speech quality of both datasets and show that Fleurs exhibits lower quality due to its data collection design. Wolof speakers were asked to read translated texts on diverse topics such as history, geography, and astronomy. However, Wolof is primarily an oral language, and its standardized script is not taught in education systems, so most speakers are not used reading it. This is reflected in the Fleurs dataset, where speech contains frequent hesitations and speakers appear less confident. We support this by comparing Fleurs to Kallaama, which consists of transcribed spontaneous speech rather than read text.
\\
\\
\myparagraph{Fleurs contains more hesitations}. We measure the Characters per second as a proxy for hesitation. We found that Fleurs' speech contains $7.51$ characters per second, over than 2 times fewer than Kallaama where speakers produce $16.88$ characters per second.

\myparagraph{Fleurs is of lower volume}. We also observed that speakers in Fleurs are less confident. This is reflected in the speech being less loud compared to Kallaama. The average volume in dB is $-49.01$ for Fleurs, while it's $-23.81$ Kallaama
\\
\\
These observations explain the speech-to-document retrieval performance on Fleurs (see Table \ref{tab:embedding_results_fleurs}). The results also suggest that, for lower speech quality, higher Matryoshka dimensions are preferable.

\subsection{To prompt or not prompt}
\begin{table*}[ht]
    \begin{minipage}[c]{.35\textwidth}
        \fontsize{9pt}{9pt}\selectfont
        %\vspace{-20.5em}
        \centering
        \addtolength{\tabcolsep}{-0.3em}
        \begin{tabular}{lcc}
            \toprule
            \textbf{Approach} & \textbf{nDCG@5} & \textbf{nDCG@10} \\
            \midrule
            Late-Fusion & \textbf{67.07} & \textbf{71.68} \\
            Dual & 46.96 & 53.70 \\
            \bottomrule
        \end{tabular}
        \caption{\textbf{Late-Fusion vs Dual when trained without prompting}. Evaluation done on Kallaama-Retrieval-Eval with dim = 1024.}
        \label{tab:wo_prompt}
    \end{minipage}
    \hfill
    \begin{minipage}[c]{.62\textwidth}
        \fontsize{9pt}{9pt}\selectfont
        \centering
        \addtolength{\tabcolsep}{-0.1em}
        \begin{tabular}{lcccc}
            \toprule
             & \multicolumn{4}{c}{\textbf{Tasks}} \\
            \cmidrule(lr){2-5}
             & \multicolumn{2}{c}{\textbf{Document Retrieval}} & \multicolumn{2}{c}{\textbf{Keyword Spotting}} \\
            \cmidrule(lr){2-3} \cmidrule(lr){4-5}
            \textbf{Prompt} & \textbf{nDCG@5} & \textbf{nDCG@10} & \textbf{F1} & \textbf{Recall} \\
            \midrule
            Document Retrieval & \cellcolor{violet!45}68.85 & \cellcolor{violet!45}74.49 & 85.35 & 87.39 \\
            Transcription Retrieval & 66.08 & 71.38 & \cellcolor{violet!45}88.79 & \cellcolor{violet!45}89.79 \\
            \bottomrule
        \end{tabular}
        \caption{\textbf{Instruction-Following performance} of \textit{Late-Fusion} (d=1024) on document retrieval and keyword spotting with different prompts.}
        \label{tab:if}
    \end{minipage}
\end{table*}
\myparagraph{Late-Fusion vs Dual without prompting.}
We ask whether the better performance of Late-Fusion compared to the Dual approach is not due to the prompting advantage of the former. We test this by training the Late-Fusion model without task prompting. We present in Table \ref{tab:wo_prompt} the results on the Kallaama-Retrieval-Eval for the Late-Fusion and Dual approaches, trained without prompting using the retrieval loss. As we can see, the Late-Fusion substantially outperforms the Dual models even without prompting. We conclude that the superiority of the Late-Fusion is not just due to the prompting capabilities.
\\
\\
\myparagraph{Instruction-Following Capabilities.}
The Late-Fusion model is multilingual, multimodal, and multitask, as the retrieval target can be specified via the prompt at inference time. We analyze the Instructon-Following capabilities of the model by testing it on retrieval and keyword-spotting performance under different prompts. Table \ref{tab:if} shows that performance is highest when the task description and task objective are aligned.

\subsection{The rank of Matryoshka dimensions}
We have seen in Section \ref{sec:unseen_task} that smaller dimensions need more training time to catch up on the higher dimensions. However, for retrieval, all the dimensions are trained on the same amount of data, and the results in Table \ref{tab:embedding_results_kallama} and \ref{tab:embedding_results_fleurs} show that small dimensions perform worst. This suggest that small dimensions fail to represent the full information. We study next the rank of the Matryoshka dimensions to better provide more substance to interpretate these results.
\\
\\
\myparagraph{Measuring the rank} There are many ways to study the rank. We analyze the cumulative energy ratio $R(k)$, which is the proportion of total variance explained by the top-k eigenvalues of the covariance matrix: $R(k) = \frac{\sum_{i=1}^{k} \lambda_i}{\sum_{j=1}^{d} \lambda_j}$ $\text{for } k \in \{1, 2, \ldots, d\}$. Where $\lambda_i$ are the eigenvalues of the covariance matrix, sorted in descending order: $\lambda_1 \geq \lambda_2 \geq \ldots \geq \lambda_d$. For each Matryoshka dimension, we compute the covariance matrix on the full test dataset of Kallaama. This ratio tells us how the information is spread within the Matryoshka vector. If the vector is low rank, then $R$ reaches $1.0$ (100\%) more quickly, which means it needs only few dimensions to represent the full information.
\\
\\
\myparagraph{Result 1: Ranks of Matryoshka dimensions.} Figure \ref{fig:energy_evolution} shows the energy distribution for different Matryoshka dimensions. Higher dimensions reach 1.0 more quickly, indicating they are lower rank and require only a small fraction of the full dimension to represent information. However, as shown in Tables \ref{tab:embedding_results_kallama} and \ref{tab:embedding_results_fleurs}, lower dimensions fail to match their performance. If higher dimensions are low-rank, why can't lower dimensions achieve similar results? This suggests that Matryoshka representation learning fails to retain critical information during compression. It has been suggested \citep{zhang-etal-2025-smec, wen2025Matryoshkarevisitingsparsecoding} that MRL's joint training and rigid compression rule, which is just a slicing, increases gradient variance during training and discards critical information.
\\
\\
\myparagraph{Result 2: Rank of queries and documents.} We also compare the rank of speech queries and documents. Figure \ref{fig:energy_100p} shows the fraction of dimensions needed to represent the full energy. Documents have higher ranks than queries since they contain more information. At dimensions 128 and 256, the vectors are full-rank, which may explain their lower performance, as all document information might not be fully represented.
\begin{tcolorbox}[compactbox,colback=violet!15!white,colframe=violet!100!white,title={\color{black} \textbf{Finding 5}: Information is inequally distributed in Matryoshka dimensions.}]
The rank analysis of the Matryoshka dimensions shows that most information is concentrated in a few dimensions, suggesting that MRL does not achieve optimal compression and leaves room for further compression.
\end{tcolorbox}
\subsection{Costs}

\begin{figure}[ht]
    \includegraphics[width=\linewidth]{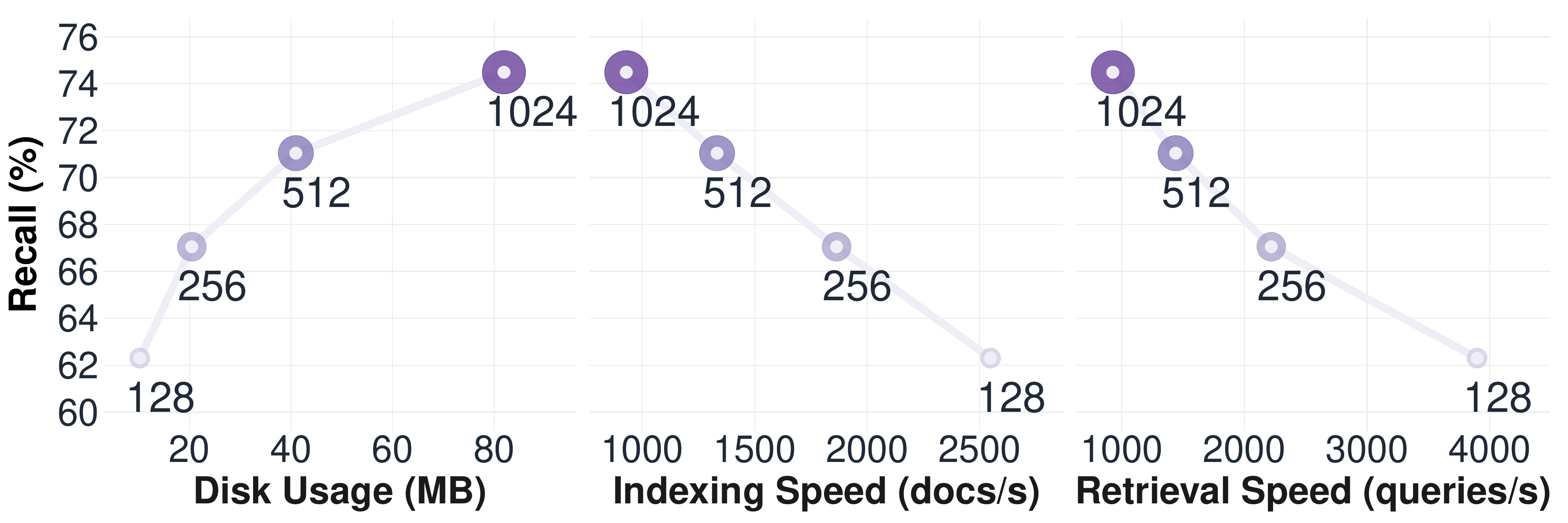}
    \caption{\textbf{Costs} across Matryoshka dimensions.}
    \label{fig:costs}
\end{figure}
\myparagraph{Method}. We analyze the Matryoshka dimensions costs using \texttt{chromadb}, which is an open-source application for search and retrieval. We embed in float16 precision the documents in the different Matryoshka dimension. We measure the time in second for indexing documents (docs/s) and the final disk usage in MB. We then measure the time to query the documents store for one result.
\\
\\
\myparagraph{Results.} Figure \ref{fig:costs} presents the relationship between performance and costs. For disk usage, we observe diminishing returns: increasing the Matryoshka dimension yields marginal accuracy gains while incurring higher memory costs. Smaller embedding dimensions enable faster retrieval, particularly at lower dimensions due to reduced FLOPs. However, indexing speed improves more slowly. The trade-offs depend on the user's compute and the downstream task. As shown in Section \ref{sec:unseen_task}, smaller dimensions can match the performance of bigger dimensions with more training data.

\section{Conclusion and Future Work}
We presented the first cross-lingual speech-text Matryoshka embedding models for Wolof and French, enabling direct retrieval of French documents from Wolof speech. We collect large-scale speech-text training data and compare modeling approaches, showing that late modality fusion within a frozen text Matryoshka model consistently outperforms dual approaches. Our rank analysis reveals that information concentrates in a few dimensions, suggesting suboptimal compression in current Matryoshka training methods.

\section{Limitations}

\myparagraph{Synthetic Dataset.} The majority of the training data relies on synthetic documents generated from transcribed and translated speech. While this enables scaling in low-resource settings, synthetic data may not fully capture the diversity and noise of real-world documents, potentially limiting robustness at deployment time. Morever, this approach supposes to have functional ASR and translation systems, which are not always available for many languages.
\\
\\
\myparagraph{Beyond Matryoshkas.} Our analysis of Matryoshka Representation Learning reveals that only few dimensions represent the information, suggesting suboptimal compression. There are many new works exploring alternatives to Matryoshka Representation Learning \citep{wen2025Matryoshkarevisitingsparsecoding}. We plan to explore dynamic structured sparsity, a more granular and flexible compression approach.
\section{Ethical Considerations}
We curated the speech dataset from multiple sources, some with restrictive licenses. However, they can still be used for research under fair-use conditions, so we plan to release both the data and the trained models under a non-commercial, research-only license.
\section{Acknowledgement}
This project was provided with computing HPC and storage resources by GENCI at IDRIS thanks to the grant 2025-AD011011668R5 and AD011014953 on the supercomputer Jean Zay’s A100 and H100 partition
\bibliography{custom}
\appendix
\section{Multitask Capabilities}
\subsection{Beyond speech-to-document retrieval}
\label{sec:asr}
We extend the synthetic speech-to-document dataset with natural ASR data from multiple sources: \textbf{FLEURS}, \textbf{ALFA} \cite{gauthierSSL}, \textbf{Common Voice} (\textbf{CV}), \textbf{Kallaama} \cite{kallama}, and \textbf{Urban-Bus} (\textbf{UB}) \cite{urbanbus}. We retain the official train/test splits for FLEURS and create random splits for the other datasets. During training, speech serves as the query and transcriptions or their translations as retrieval targets. Table \ref{tab:asr_data} summarizes the dataset distribution.
\begin{table}[ht]
    \centering
    {\small{
    \begin{tabular}{lccccc}
        \toprule
        \textbf{\textit{Split}} & \textbf{Fleurs} & \textbf{Alfa} & \textbf{CV} & \textbf{Kallaama} & \textbf{UB} \\
        \midrule
        \textbf{\textit{Train}} & 8.72 & 16.13 & 34.97 & 33.60 & 4.52 \\
        \textbf{\textit{Test}} & 1.75 & 2.84 & 6.21 & 5.91 & 1.12 \\
        \bottomrule
    \end{tabular}
    }}
\caption{The non-synthetic ASR dataset.}
    \label{tab:asr_data}
\end{table}

\end{document}